# Autonomous Assessment of Demonstration Sufficiency via Bayesian Inverse Reinforcement Learning

Tu Trinh
Center for Human-Compatible
Artificial Intelligence,
University of California, Berkeley
Berkeley, CA, USA
tutrinh@berkeley.edu

Haoyu Chen
Department of Mathematics
University of Utah
Salt Lake City, UT, USA
haoyu.chen@utah.edu

Daniel S. Brown
Kahlert School of Computing
University of Utah
Salt Lake City, UT, USA
daniel.s.brown@utah.edu

## ABSTRACT

We examine the problem of determining demonstration sufficiency: how can a robot self-assess whether it has received enough demonstrations from an expert to ensure a desired level of performance? To address this problem, we propose a novel self-assessment approach based on Bayesian inverse reinforcement learning and value-at-risk, enabling learning-from-demonstration ("LfD") robots to compute high-confidence bounds on their performance and use these bounds to determine when they have a sufficient number of demonstrations. We propose and evaluate two definitions of sufficiency: (1) normalized expected value difference, which measures regret with respect to the human's unobserved reward function, and (2) percent improvement over a baseline policy. We demonstrate how to formulate high-confidence bounds on both of these metrics. We evaluate our approach in simulation for both discrete and continuous state-space domains and illustrate the feasibility of developing a robotic system that can accurately evaluate demonstration sufficiency. We also show that the robot can utilize active learning in asking for demonstrations from specific states which results in fewer demos needed for the robot to still maintain high confidence in its policy. Finally, via a user study, we show that our approach successfully enables robots to perform at users' desired performance levels, without needing too many or perfectly optimal demonstrations.

## 1 INTRODUCTION

If robots and other AI systems are to be deployed in safety-critical settings, we want them to be able to confidently self-assess their performance and understand when they require additional training data to meet a desired level of performance. This becomes especially difficult in domains such as learning from demonstration (LfD) [3, 39] or inverse reinforcement learning (IRL) [47, 4] where the reward function itself is unknown and must be learned from human demonstrations. Imagine a robot that is learning from demonstrations how to navigate across uneven terrain or how to perform a household task such as putting dishes away. How many demonstrations are needed for the robot to learn the task? Is there a way for the robot to self-assess whether it has received a sufficient number of demonstrations needed to achieve good performance? How can the robot measure performance if it doesn't know the human's true reward function? In this paper we seek to address these questions.

Our main insight is the following: *Maintaining a belief distribution over the demonstrator's true, but unobserved, reward function, enables a robot to reason about its performance under this distribution and determine, with high-confidence, when it has received enough demonstrations to satisfy a desired performance threshold.*

To maintain a belief distribution over reward functions, we propose a novel application of Bayesian IRL (BIRL) [38] that uses samples from the posterior distribution over reward functions given demonstrations to enable the robot to evaluate its current policy and determine how confident it is that this learned policy has sufficiently good performance. We propose two definitions of demonstration sufficiency: (1) whether, with high confidence, the learned policy has low regret compared to the optimal policy under the unobserved reward function of the demonstrator and (2) whether the learned policy will, with high confidence, outperform a given baseline policy (e.g., a policy that is known to be safe but suboptimal) by a desired margin. Our approach allows a robot to self-assess when it has received enough demonstrations to enable it to meet one of the above performance criteria.

By proposing a Bayesian approach to self-assessment when learning from demonstrations, we seek to enable robots to properly reason about uncertainty. For example, if the human demonstrator happens to provide redundant or ambiguous demonstrations, the robot will have a large amount of uncertainty regarding the true intention of the demonstrator, leading it to continue to ask for additional demonstrations. This kind of self-assessment also allows robots to know when they do not require any further demonstrations. An important benefit of this approach is that it removes the need for the human to predict when the robot has had enough training data. It is often difficult for humans to inspect a robot's policy or learned reward function to determine whether it is aligned with their intent. Instead, we argue that robots should be able to self-assess their performance, relative to their uncertainty over the human's intent. Figure 1 shows an illustrative example of this.

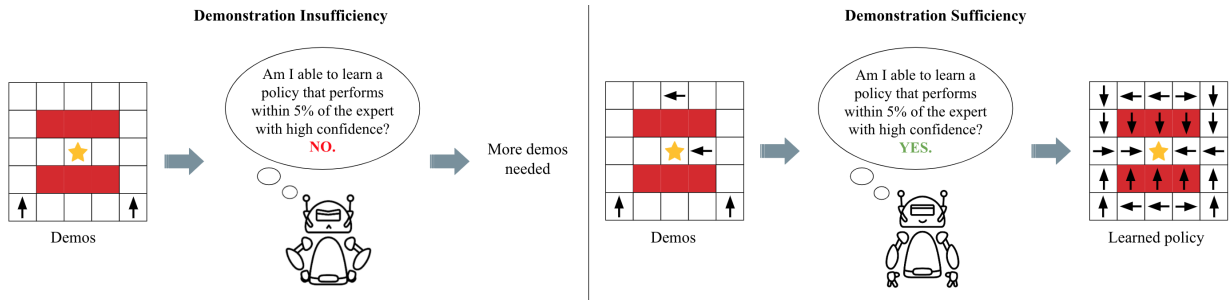

**Figure 1:** *Demonstration sufficiency:* Pictured is an illustrative two-feature MDP in which the demonstrator is seeking to teach the robot to reach the starred state. White states have small negative reward and red states have large negative reward. (Left) The robot receives two demonstrations (the black arrows) from the expert but, through our self-assessment method discussed below, deems them insufficient to be highly confident that its learned policy has low regret–it does not yet have strong evidence about the relative rewards of the different features in the MDP. (Right) The robot receives an additional two demonstrations and, after applying our method, deems them sufficient to guarantee with high-confidence that its learned policy will have low regret if evaluated under the unobserved true reward function.

We examine demonstration sufficiency assessment across several domains using both simulated demonstrations as well as human-provided demonstrations from a user study. The main contributions of our work are: (1) we formalize the problem of demonstration sufficiency for robots that learn from demonstrations; (2) we define two specific metrics for measuring demonstration sufficiency and develop approaches to calculate high-confidence performance bounds on these metrics through a novel application of Bayesian inverse reinforcement learning; and (3) we evaluate our approach across four domains in simulation and two domains through a user study, showing the efficacy of our approach in enabling a robot to accurately assess when it has received enough demonstrations.

## 2 RELATED WORK

Our work falls under the area of autonomous self-assessment of robots and other AI systems. Previous work examines how a robot can assess its performance and communicate its shortcomings to a human expert [36, 17]; however, most existing performance metrics and studies do not involve learning from demonstration, and those that do, focus on communication and knowledge-sharing [29, 24] rather than addressing how a robot or AI agent can directly self-assess whether a learned policy or reward function is above a desired safety threshold. Other work [7] studies optimal stopping for robot teaching but uses information gain from pairwise preferences instead of policy performance estimated from demonstrations. Prior work does consider high-confidence performance bounds for inverse reinforcement learning [1, 42]. However, the bounds obtained by these methods are generally loose and correlate to a high number of training examples needed to show the robot. We build off more recent work [14, 15, 13] that demonstrate tighter bounds on performance but do not consider how these bounds can be used for autonomous assessment of demonstration sufficiency.

Our work is also related to prior work on pedagogic teaching by demonstrations which studies how to craft demonstrations that will be maximally informative [18, 16, 28, 45, 30]; however, prior work only considers this problem from the teacher's perspective and assumes the teacher has privileged information about the student's learning algorithm and complete knowledge of the reward function they seek to teach. By contrast, we focus on developing algorithms from the student's perspective, i.e., algorithms that allow the robot to know when it has received sufficient demonstrations, without any assumptions about the demonstrations being highly informative.

## 3 METHOD

### 3.1 Preliminaries

*3.1.1 Markov Decision Processes.* We model the environment as an MDP consisting of a set of states $S$, actions $A$, transition function $T : S \times A \times S \rightarrow [0, 1]$, reward function $R : S \rightarrow \mathbb{R}$, initial state distribution $S_0$, and discount factor $\gamma \in [0, 1)$. A policy $\pi$ is a mapping from states to a probability distribution over actions. The expected return of a policy $\pi$ under a reward function $R$ is denoted as $V_R^\pi = \mathbb{E}_{s \sim S_0} V_R^\pi(s)$, where the value function for a state is $V_R^\pi(s) = \mathbb{E}_\pi [\sum_{t=0}^\infty \gamma_t R(s_t) | s_0 = s]$ and the Q-value function for a state-action pair is defined as $Q_R^\pi(s, a) = R(s) + \gamma \sum_{s' \in S} T(s, a, s) V_R^\pi(s')$. Following prior work [1, 47, 14, 26, 4], we assume that the reward function $R$ can be defined in terms of a linear combination of features: for an MDP with features $\phi(s) \in \mathbb{R}^k$, $R(s) = w^T \phi(s)$ where $w \in \mathbb{R}^k$ is a vector of feature weights, with $\|w\|_2 = 1$.

*3.1.2 Bayesian Inverse Reinforcement Learning.* In inverse reinforcement learning (IRL), we seek the underlying reward function of an MDP given demonstrations [35]. We denote a set of demonstrations by $D$, which we define to be a set of state-action pairs: $D = \{(s_1, a_1), \ldots, (s_n, a_n)\}$. Bayesian IRL (BIRL) [38] estimates the posterior distribution over reward functions given demonstrations, $P(R|D) \propto P(D|R)P(R)$, where the demonstrator is assumed to follow a softmax policy, leading to the following likelihood function:

$$P(D|R) = \prod_{(s,a) \in D} P((s,a)|R) = \prod_{(s,a) \in D} \frac{e^{\beta Q_R^*(s,a)}}{\sum_{b \in A} e^{\beta Q_R^*(s,b)}} \quad (1)$$

where $\beta \in [0, \infty)$ represents the confidence in the demonstrator's optimality (a higher $\beta$ means the demonstrator is more likely to give optimal demonstrations) and $Q_R^*(s,a) = \max_\pi Q_R^\pi(s,a)$ is the optimal Q-value for a state and action under the reward function $R$. Equation (1) assigns higher likelihoods to demonstrated actions that result in higher Q-values under $R$ compared to alternative actions. Equation (1) is an example of Boltzmann-rationality, a model that has found widespread utility in economics [11, 33], psychology [5, 22, 23], and AI [48, 19, 8, 20, 25, 31] as a useful model of human decision-making and can be seen as the maximum entropy distribution over choices for a satisficing agent [25].

*3.1.3 Value-at-Risk Bounds.* Value-at-Risk is a common probabilistic measure of worst-case performance [27, 43]. The $\alpha$-Value-at-Risk, or $\alpha$-VaR, is the $\alpha$-worst-case value of a random variable $Z$. With $\alpha \in (0,1)$ as the quantile level, this is defined as

$$v_\alpha(Z) = F_Z^{-1}(\alpha) = \inf\{z : F_Z(z) \geq \alpha\} \quad (2)$$

where $F_Z(z) = P(Z \leq z)$, the cumulative distribution function of $Z$. The higher the value of $\alpha$, the more risk-sensitive we are.

## 3.2 Problem Definition

Our aim is to determine whether or not a robot has received sufficient demonstrations in order to complete a task in a way that aligns with the expert's intended policy derived from their unobserved reward function $R^*$. We want to enable robots to quantify how confident they are in the goodness of their policy compared to the expert's via high-confidence bounds on regret. The robot should request more demonstrations if it is not yet highly confident that its learned policy will have low regret compared to the expert's, and it should declare demonstration sufficiency if it is highly confident.

**Demonstration Sufficiency:** Given an MDP with an unobserved reward function $R^*$, a set of demonstrations $D$, confidence parameter $\alpha$, and a performance threshold $\varepsilon$, we want the robot to be able to determine when it is $\alpha$-confident that its policy regret, if evaluated under the demonstrator's true reward function $R^*$, is no worse than $\varepsilon$. Thus, demonstration sufficiency is achieved when

$$P\left(regret(\pi_{\text{robot}}, R^*) \leq \varepsilon \mid D\right) \geq \alpha \quad (3)$$

Given this self-assessment, the robot can tell the demonstrator that it has received enough demonstrations. Note that in practice, $\pi_{\text{robot}}$ can be any robot policy. In our paper, we set $\pi_{\text{robot}}$ to be $\pi_{\text{MAP}}$, the optimal policy corresponding to the maximum a posteriori reward estimate $R_{\text{MAP}}$ learned by the robot using Bayesian IRL.

## 3.3 Determining Demonstration Sufficiency

To assess demonstration sufficiency we must select a measure of regret. We also need to figure out how the robot can bound its performance with respect to the unknown reward function, $R^*$.

To represent policy regret (also known as "policy loss"), prior work on IRL [38, 32, 44, 14, 15] has typically used the expected value difference (EVD), defined as

$$EVD(\pi_{\text{robot}}, R^*) = V_{R^*}^* - V_{R^*}^{\pi_{\text{robot}}} \quad (4)$$

While this is a common metric for comparing different IRL algorithms, reward functions are equivalent under positive scaling and affine shifts [35, 2], making a threshold defined in raw reward units uninterpretable. Thus, we propose the use of demonstration sufficiency thresholds defined in terms of normalized expected value difference (nEVD):

$$regret(\pi_{\text{robot}}, R^*) := nEVD(\pi_{\text{robot}}, R^*) = \frac{V_{R^*}^* - V_{R^*}^{\pi_{\text{robot}}}}{V_{R^*}^* - V_{R^*}^{\pi_{\text{rand}}}}, \quad (5)$$

where $\pi_{\text{rand}}$ is a uniform random policy. Normalizing with respect to a random uniform policy enables the demonstrator to specify a regret threshold in terms of an interpretable percentage rather than in raw reward units.

Because the ground-truth reward function is unknown to the robot, it is impossible to calculate its true regret, $nEVD(\pi_{\text{robot}}, R^*)$. Instead, to perform demonstration sufficiency self-assessment, we propose a Bayesian approach that leverages Bayesian IRL to sample from $P(R|D)$, the posterior distribution of reward functions given demonstrations, and then uses these reward samples to calculate an $\alpha$-Value-at-Risk ($\alpha$-VaR) upper bound on regret [14, 15]. We propose that the robot should declare demonstration sufficiency when

$$v_\alpha\left(nEVD(\pi_{\text{robot}}, R)\right) \leq \varepsilon, \text{ for } R \sim P(R|D). \quad (6)$$

## 3.4 Dealing with Finite Sampling Errors

As discussed in the previous section, we want to find an $\alpha$-quantile worst-case bound on $nEVD(\pi_{\text{robot}}, R^*)$ by computing the $\alpha$-VaR over $P(R|D)$. In practice we do not know $P(R|D)$ explicitly but instead obtain samples from the posterior $P(R|D)$ via Markov chain Monte Carlo (MCMC) methods [38]. Thus, we need to be careful about the error induced by samples and make sure that we do not underestimate the policy regret due to a small number of samples from the posterior. To find $P(R|D)$, we use MCMC to generate a sequence of sampled rewards $\mathcal{R} = \{R : R \sim P(R|D)\}$ from the posterior distribution over reward functions given the demonstrations. Bayesian IRL has rapid finite-time mixing guarantees and converges to the true posterior, making it a viable method to estimate $P(R|D)$ [38], but we still need to deal with error and uncertainty when estimating the value-at-risk. We do this as follows. For each sample $R_i \sim P(R|D) \in \mathcal{R}$ we first compute

$$X_i = nEVD(\pi_{\text{robot}}, R_i), \quad (7)$$

giving us samples from the posterior distribution over normalized expected value differences conditioned on the user's demonstrations. Given $n$ samples of $X$, we can obtain a point estimate of the $\alpha$-VaR by sorting the samples of $X$ in ascending order to get order statistics $Z$, then take the $\alpha$-quantile. This gives us $Z_k$ as an estimate of the $\alpha$-VaR, where $k = \lceil \alpha n \rceil$. However, this does not take into account our confidence in this point estimate.

Following Brown et al. [14], we derive a high-confidence upper bound on the $\alpha$-VaR. By definition, we have that $P(X_i < v_\alpha(X)) = \alpha$ for any sample $X_i, i \in 0, \ldots, n$. We first sort these samples to obtain order statistics $Z_j, j \in 0, \ldots, n$. Then for any $Z_j$, we can calculate the probability that the $\alpha$-VaR is less than $Z_j$ using the binomial cumulative distribution function (CDF):

$$P(v_\alpha(X) < Z_j) = F(j-1; n, \alpha) \quad (8)$$
$$= \sum_{i=0}^{j-1} \binom{n}{i} \alpha^i (1-\alpha)^{n-i} \quad (9)$$

Note that $v_\alpha(X)$ is the $100\alpha$ percentile value of $X$. Thus, for the order statistic $Z_j$ to be larger than $v_\alpha(X)$, we must have that $v_\alpha(X)$ is greater than at most $j-1$ samples. This probability is given by the binomial CDF, $F(j - 1; n, \alpha)$, which gives the probability of getting $j - 1$ or fewer successes in $n$ trials (hence Eq. (8)). In this formulation, a success is when a sample $X_i$ is less than $v_\alpha(X)$, making the probability of success, $P(X_i < v_\alpha(X))$, equal to $\alpha$ by definition of $\alpha$-VaR; it follows that the probability of failure, $P(X_i \geq v_\alpha(X))$, is $1 - \alpha$, hence Eq. (9). Finally, to get a 95% confidence bound on $v_\alpha(X)$, we can use the inverse binomial CDF, $F^{-1}$. Thus, the order statistic $Z_k$, where $k = F^{-1}(0.95; n, \alpha)$, forms a 0.95-confidence bound on $v_\alpha(X)$. We use the above derivation to compute 95%-confidence bounds on the $\alpha$-VaR throughout the paper.

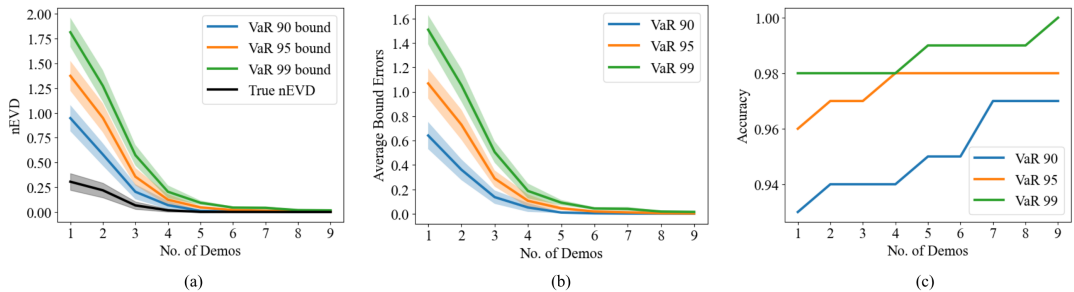

Figure 2: *Bounding regret with $\alpha$-VaR:* Error bands around the lines denote standard error. (a) shows that the robot is able to generate accurate $\alpha$-VaR upper bounds on its true, but unobserved, regret (nEVD). (b) shows the corresponding bound errors and (c) shows the accuracy, indicating that in almost all MDPs, the bounds are correct upper bounds.

### 3.5 Empirical Justification for Our Bounds

Figure 2 shows the results of testing the robot's capability of estimating the underlying reward function using our approach to bound the nEVD in a simple 5x5, 4-feature gridworld environment. We set the confidence level $\alpha$ to be 0.90, 0.95, and 0.99 and plotted the results of each trial. The $x$-axis denotes the number of demonstrations the robot is given. We show the actual bounds, the bound error (how much the estimated policy loss overestimates the true policy loss on the unobserved ground-truth reward), and accuracy, defined as the proportion of environment replicates in which the $\alpha$-VaR bound on the nEVD is higher than the true nEVD.

We see that the robot is able to obtain high accuracy and tight value-at-risk (VaR) bounds on the normalized expected value difference. Starting with only one or two demos, the robot cannot be expected to truly learn what is an acceptable policy or reward function, and the high bounds and bound errors reflect this fact. However, within five or six demos, the VaR bounds and the bound errors are all able to reach close to 0.0. Our results provide strong evidence that we can use our previously described approaches to both bound performance and to determine demonstration sufficiency. As expected, the higher $\alpha$ is, the more conservative the bound; however, all bounds are highly accurate and surprisingly tight, especially for increasing numbers of demonstrations.

## 4 EMPIRICAL RESULTS
### 4.1 Experimental Design
Figure 3 shows the environments we use to test our methodology. Two have discrete state spaces (Gridworld and Driving) and two have continuous state spaces (Lunar Lander and Lavaworld).

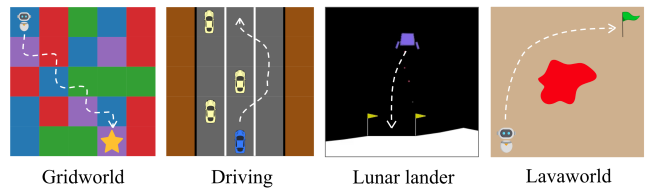

Figure 3: The environments in which we examine demonstration sufficiency assessment.

We generated multiple randomized MDP instances of each environment and tested three methods on the same set of MDPs: our approach and two baselines, discussed in Section 4.2. In each environment we simulate human demonstrations by sampling states uniformly at random and providing an optimal action for that state (discrete environments) or optimal trajectory starting from that state (continuous environments). Our test environments are:

**Gridworld:** A discrete state space MxN environment where each state has one of four features, each associated with a different reward weight. One of these states is the goal state. The robot can take one of four actions: up, down, left, and right.

**Driving:** A discrete, infinite horizon environment with different road conditions and traffic. There are two off-road patches on either side of the main roads. The end of the road segment is connected to the beginning, simulating a continuous road. Actions are drive straight, turn left, and turn right.

**Lunar lander [12]:** A continuous state space environment from OpenAI Gym where the human lands a craft on the moon.

**Lavaworld [26]:** A continuous state space environment where the human guides the robot towards a goal while attempting to both avoid a pit of lava randomly placed in the environment and maintain a smooth, non-jerky trajectory.

### 4.2 Baselines
To the best of our knowledge, we are the first to study demonstration sufficiency. Thus, we adapt two stopping criteria from supervised learning [37] into heuristic baselines for demonstration sufficiency.

**Convergence (Conv.)**: Given a "patience" hyperparameter $p$, the robot signals demonstration sufficiency when $\pi_{\text{MAP}}$ does not change over $p$ consecutive demonstrations.

**Validation set (V.S.)**: Every $i^{\text{th}}$ demonstration is added to a held-out set. If for each $(s, a)$ in the held-out set, $\pi_{\text{MAP}}(s) = a$, the robot declares demonstration sufficiency.

### 4.3 Dependent Measures

**Identification accuracy:** We use *F1 score* to represent identification accuracy, defined as

$$F1 = \frac{TP}{TP + \frac{1}{2}(FP + FN)} \quad (10)$$

True positive (TP) means that $regret(\pi_{\text{robot}}, R^*) \leq \varepsilon$ when the robot declares demonstration sufficiency. False positive (FP) is when the robot declares demonstration sufficiency but $regret(\pi_{\text{robot}}, R^*) > \varepsilon$. False negative (FN) is when the robot does not declare demonstration sufficiency but $regret(\pi_{\text{robot}}, R^*)$ is actually less than $\varepsilon$.

**Sample efficiency:** While having good accuracy is important, we argue that practicality in terms of human interaction burden is also crucial. More specifically, we do not want the human to have to give too many demonstrations to the robot before it can learn a high-performing policy. As such, we measure the *proportion of demonstrations needed* before the robot determines it can stop receiving demonstrations. For discrete environments this is the number of unique states where a demonstrator action is provided. For continuous environments we use the number of demonstrated trajectories.

### 4.4 Analysis

For our nEVD stopping condition, we tested five different thresholds: 0.1, 0.2, 0.3, 0.4, 0.5. We stop at 0.5 as this denotes a regret that is exactly half that of a random policy; any larger regret is deemed unreasonable. For the convergence baseline, we tested five different patience hyperparameters, $p = 1, 2, 3, 4, 5$. For the validation set baseline, we also tested five different interval hyperparameters, $i = 3, 4, 5, 6, 7$. We picked the two hyperparameter settings that gave the best sample efficiency vs. identification accuracy tradeoff in order to compare with our regret confidence bounding method. We set a confidence level of $\alpha = 0.95$ for our method.

We test the following hypotheses:

**H1.** Our method achieves higher F1 scores than baseline methods.
**H2.** Our method requires fewer demonstrations to be given to the robot before it declares demonstration sufficiency, compared to baseline methods.

The results in Figure 4 show that the nEVD bounding method generally outperforms both baseline stopping conditions. Complete results can be found in the appendix. In the discrete domains, our method achieves a higher F1 score, near 1.00 for all thresholds, than the validation set baseline (V.S.), while requiring at least 25% fewer demonstrations. This can be attributed to the fact that V.S. needs to set aside usable demonstrations for its held-out set, and on top of that requires an exact match between $\pi_{\text{MAP}}$ states and held-out states. The convergence baseline (Conv.) has high sample efficiency, but this comes at the cost of a much lower F1 score, a consistent trend across both the discrete and continuous domains. This can be attributed to the fact that Conv. depends on the stability of $\pi_{\text{MAP}}$, not its actual performance.

In the continuous domains (Figure 4(c) and (d)), our sample efficiency over the baseline methods becomes much clearer. We believe the difference is so stark because in environments with a continuous state space, there is more ambiguity regarding the demonstrator's true reward $R^*$. This ambiguity causes $R_{\text{MAP}}$ to vary widely, which means that the baselines end up requesting many demonstrations. This results in V.S. achieving a high F1-score due to exact matching with the optimal policy, but it comes with the aforementioned sacrifice in sample efficiency.

Our method maintains high F1 scores *and* high sample efficiency in both domains because it takes into account how well the robot's current policy $\pi_{\text{robot}}$ will perform under the ground-truth reward function compared to an expert using our high-confidence bounds. It does not require that $\pi_{\text{robot}}$ converges to or match any singular policy so long as the robot is confident that $\pi_{\text{robot}}$ achieves low regret. Moreover, a converged policy does not necessarily mean it will generalize well to the expert's true intended reward function. It may just mean that consecutive demonstrations convey very similar information. Determining demonstration sufficiency based on high-confidence bounds on nEVD not only enables robots to learn high-performing policies efficiently and accurately, but also allows human demonstrators to calibrate the robot's performance according to desired confidence levels and performance thresholds.

*4.4.1 Statistical Tests for H1.* Since the distribution of F1 scores across all methods and their corresponding thresholds did not meet normality or variance homogeneity assumptions, we conducted a Kruskal-Wallis test. Kruskal-Wallis yielded statistically significant results for both the discrete ($H = 71, p \approx 0$) and continuous ($H = 83, p \approx 0$) domains. Subsequent Dunn post-hoc tests with the Bonferroni correction and median comparisons revealed that there was a statistically significant difference between our method's F1 scores and Conv.'s ($p \approx 0$), but no significant difference between our method's scores and V.S.'s. ($p \approx 1$). Thus, our results partially support **H1**.

*4.4.2 Statistical Tests for H2.* We ran the same set of statistical tests to compare each method's sample efficiency across thresholds. Kruskal-Wallis yielded statistically significant results for both the discrete domain ($H = 358, p \approx 0$) and continuous domain ($H = 601, p \approx 0$). Dunn and median comparisons revealed that our method required fewer demonstrations for the continuous domain ($p \approx 0$) compared to the convergence baseline and fewer demonstrations for both domains ($p \approx 0$ for both) compared to the validation set baseline. While Conv. required fewer demonstrations for the discrete domain than our method, the median difference was only 12%. Our results provide strong evidence for **H2**, especially since most environments robots encounter in the real world will have continuous state spaces.

*4.4.3 Comparison to Prior Theoretical Bounds.* We next compare the efficiency at which our method obtains confidence bounds compared to prior work [1, 42] in IRL that uses Hoeffding bounds. While these works were also focused on determining the optimal number of demonstrations to achieve a policy regret bound, their bounds depend on loose concentration inequalities regarding the demonstrator's state occupancy frequencies. As a result, these bounds are highly impractical for determining real-world demonstration sufficiency. To showcase this, we averaged results across a common set of gridworld MDPs with $\alpha = 0.95$ for our method and a 95% confidence level for these other methods. Table 1 shows how many demonstrations the latter require to reach each of our policy loss thresholds. When compared with Figure 4, the results in Table 1 show that our approach provides a dramatic improvement in practicality over prior high-confidence bounds for robots that learn from demonstrations.

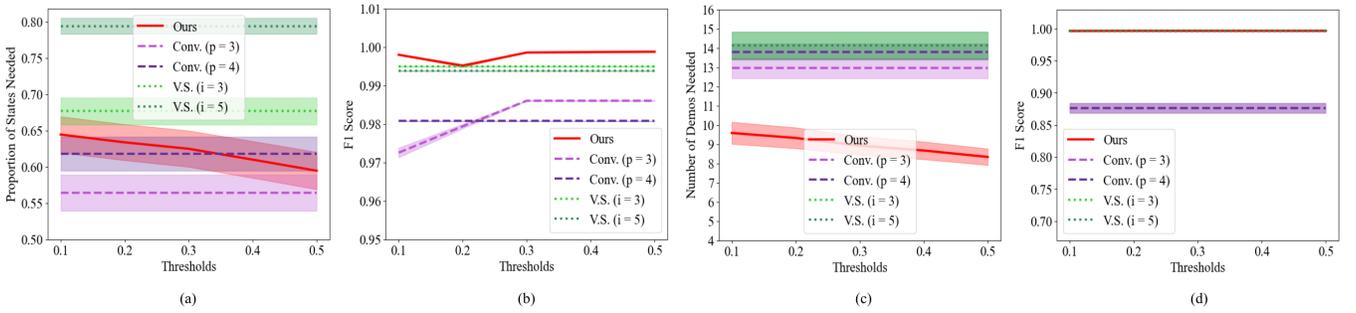

**Figure 4:** *nEVD method compared to baseline methods for determining demonstration sufficiency:* The $x$-axis across all subfigures denote the nEVD bound threshold the robot was using to assess demonstration sufficiency with our method. "Conv." denotes the convergence baseline with a patience hyperparameter of $p$. "V.S." is the validation set baseline with an interval hyperparameter of $i$. Subfigures (a) and (b) show the sample efficiency and identification accuracy measures, respectively, for the discrete domain (gridworld and driving). Subfigures (c) and (d) show the sample efficiency and identification accuracy measures, respectively, for the continuous domain (lander and lavaworld). Bands around the lines denote standard error.

| Threshold | Abbeel and Ng [1] | Syed and Schapire [42] |
| --- | --- | --- |
| 0.1 | 1,624,056 | 3,654,126 |
| 0.2 | 406,014 | 913,532 |
| 0.3 | 180,451 | 406,014 |
| 0.4 | 101,504 | 228,383 |
| 0.5 | 64,963 | 146,166 |

**Table 1:** Number of demonstrations required under a 95% confidence Hoeffding-based bound to reach each nEVD threshold. Our method (showcased in Figure 4) requires orders of magnitude fewer demonstrations to reach the same bounds.

*4.4.4 Noisy Demonstrations Ablation.* Finally, we study how our method performs given noisy, or suboptimal, demonstrations. We ran a small experiment that varied the percentage of noisy demonstrations and assessed how identification accuracy and sample efficiency changed as noise increased. On average, we found that identification accuracy decreases slowly with noise, remaining above 95% until over around 30% of demonstrations are suboptimal. The same trend could be found for the true positive rate, indicating that even with noisy demonstrations, the robot using our nEVD bounds is still able to correctly pinpoint at which point it can safely stop receiving training data. On the other hand, the false positive rate increases faster but still remains below 5% until around 20% of demonstrations are suboptimal. This trend is expected since the robot will be misled towards an incorrect reward and policy given very noisy demonstrations. Meanwhile, we found that there was no clear trend in relation to noise when it came to sample efficiency; across all noise levels, sample efficiency datapoints remained roughly within ±6% of each other. Overall, this experiment provides some evidence that our methods are decently robust to noise—20% to 30% of demonstrations can be suboptimal, which is promising for real-world applications.

### 4.5 Method Extensions

*4.5.1 Percent Improvement over a Baseline Policy.* As mentioned in the Introduction, our framework of using high-confidence bounds to help robots reason under uncertainty regarding their policy performance can be applied to another flavor of demonstration sufficiency, one based on performance gain rather than loss.

There are situations where there already exists a baseline policy, e.g., a robot comes pre-deployed with a default policy or the demonstrator has previously trained a safe policy for one task and now wants to teach the robot a related task. In such scenarios, a stopping condition based on bounds on improvement over the baseline policy would allow the robot to, with high confidence, learn a policy that performs better under the true reward function. We define this as Percent Improvement Over a Baseline (PIOB):

$$PIOB(\pi_{\text{robot}}, \pi_{\text{base}}, R) = \frac{V_R^{\pi_{\text{robot}}} - V_R^{\pi_{\text{base}}}}{V_R^{\pi_{\text{base}}}} \quad (11)$$

Using the same approach as before, we sample reward functions from the Bayesian posterior given demonstrations and use these samples to create a bound on performance gain at a given confidence level. The robot signals demonstration sufficiency when its estimated lower bound on PIOB surpasses the user-provided improvement threshold. Since the robot is trying to obtain a lower bound on policy improvement rather than an upper bound on policy loss, the robot uses a $(1 - \alpha)$-worst-case value:

$$v_{1-\alpha}(Z) = F_Z^{-1}(1 - \alpha) = \sup\{z : F_Z(z) \leq (1 - \alpha)\} \quad (12)$$

Given a set of demonstrations $D$, a baseline policy $\pi_{\text{base}}$, and an improvement threshold $\varepsilon$, demonstration sufficiency is now determined by whether the robot policy sufficiently improves over the baseline with high confidence.

| PIOB Bound Threshold | Discrete F1 Score | Continuous F1 Score |
|---|---|---|
| 20% | 1.00 ± 0.00 | 0.99 ± 0.00 |
| 40% | 0.97 ± 0.00 | 0.99 ± 0.00 |
| 60% | 0.95 ± 0.00 | 0.99 ± 0.00 |

Table 2: *PIOB method:* F1 scores for the two domains, for each percent improvement bound threshold used.

| Stopping Condition | Reduction Mean | StdDev |
|---|---|---|
| nEVD | 12.95% | 1.62% |
| Percent improvement | 13.24% | 0.65% |

Table 3: *Passive vs. active demonstration selection:* The mean and standard deviation of reduction in proportion of states needed between active and passive demonstration selection.

Using the same four environments, we show the results of using VaR bounds on percent improvement over a baseline as a way to assess demonstration sufficiency in Table 2 (more results can be found in the appendix). We omitted the sample efficiency measure from the table due to space constraints; we found that sample efficiency was similar to what was achieved with the nEVD bound method (67% ± 2% of states for discrete, 6.12 ± 0.22 states for continuous) and, as expected, decreased with increasing threshold values, especially if the original baseline policy was already high-performing. This trend can also be seen in the F1 scores. While these scores start out high, similar to those achieved with nEVD bounds, they decrease as the threshold increases because the robot accrues more false negatives—due to the conservative nature of our high-confidence performance bounds. We do not see this as a large concern because these bounds are designed to be lower bounds on policy gain. That is, the robot may underestimate the quality of its learned policy, which in reality can turn out to be better than expected.

*4.5.2 Active Learning.* Our previous empirical experiments and user study used demonstrations that were given in a "passive" manner, where demonstrations were chosen by the human. In this section, we investigate the benefits of applying prior work on risk-aware active queries [15] to allow the robot to actively query for demonstrations, which can achieve better sample efficiency while still maintaining high demonstration sufficiency identification accuracy.

For the two discrete environments where there are a finite set of states to select from, we conducted experiments using the same MDPs and dependent measures for both nEVD and PIOB bounding methods, with a key difference being that the robot is able to actively query a specific state where it wants an additional demonstration. At each iteration, the robot calculates which state has the highest $\alpha$-VaR bound on expected value difference (EVD), then requests a demonstration from this state to be added to the demo set $D$ for the next iteration. Note that we use unnormalized state EVD (similar to Eq. (4) but for a single state instead of the whole policy). This is because the normalization factor helps quantify the performance of a policy but is unnecessary computation if we are merely comparing the EVDs of different states together. Formally, the robot will select a state, $s^*$, for requesting a new demonstration as follows:

$$s^* = \underset{s \in S}{\operatorname{argmax}}\, \nu_\alpha(EVD(s, R)) = \underset{s \in S}{\operatorname{argmax}}\, \nu_\alpha(V_R^*(s) - V_R^{\pi_{\text{robot}}}(s)) \quad (13)$$

where $R$ are the reward functions sampled from the Bayesian posterior and $V_R^*(s)$ and $V_R^{\pi_{\text{robot}}}(s)$ are the values of state $s$ under $R$ for the respective policies.

Note that both the nEVD bound and PIOB bound stopping conditions use EVD in selecting a state to actively query. While in practice the demonstrator can set the selection metric to be any measure he or she prefers, we believe that it is best for the robot to select states based on which one currently results in the most policy regret compared to an expert to ensure as high-performing a policy as possible.

We find that active demonstration selection results in significantly fewer state-action pairs required for the robot to signal demonstration sufficiency, compared to passive demonstration selection, with no compromise in identification accuracy. When actively querying, the robot is able to pinpoint exactly what information it needs before it can be confident in learning a high-performing policy, instead of the demonstrator having to guess what information would be most useful. This reduction in the percentage of states needed is showcased in Table 3.

## 5 USER STUDY

We designed a user study in order to evaluate our approach with real human demonstrations. We recruited 11 participants from the university campus, aged 18-55, 64% male, 36% female.

### 5.1 Experimental Design and Measures

To keep the user study session within a reasonable amount of time, we designed six experiments for our participants to execute via an online interface: two environments × three demonstration sufficiency methods. We chose the gridworld and driving environments, which would be both easy enough for users to handle during the study, and, like in the real world, difficult enough where users could provide good, but not always optimal, demonstrations [34]. The three methods used were nEVD bounds (ours), convergence, and validation set. We used $\alpha = 0.95$ and an nEVD bound threshold of 0.3 for our method, $p = 3$ for convergence, and $i = 5$ for validation set (the median hyperparameter values for each method). For each experiment, participants were either given a randomly generated reward function or asked to design their own for the environment. Then, they provided demonstrations one by one for the robot. After the robot declared demonstration sufficiency, users were asked to rank the final learned policy on a scale of 1 to 5 based on how well they thought the policy aligned with their demonstrations and intended reward function. This study was blind in that the participants were never aware which of the three methodologies was operating behind the scenes. A more detailed description of the study can be found in the appendix.

| Evaluation Metric | Gridworld | | | Driving | | |
|---|---|---|---|---|---|---|
| | Ours | Convergence | Validation Set | Ours | Convergence | Validation Set |
| Prop. states needed | 0.22 ± 0.05 | 0.45 ± 0.03 | 0.64 ± 0.01 | 0.25 ± 0.06 | 0.69 ± 0.07 | 0.60 ± 0.03 |
| User evaluation | 4.31 ± 0.40 | 3.47 ± 0.32 | 4.00 ± 0.53 | 4.15 ± 0.28 | 3.00 ± 0.71 | 4.00 ± 0.39 |

Table 4: *User study results:* The three methods' performances in our user study.

## 5.2 Analysis

We test the following hypotheses:

**H3.** Users liked the policies that our method learned more than those that the baseline methods learned.

**H4.** The proportion of demonstrated states our method required in the user study is less than what the baseline methods required.

Results from our user study are shown in Table 4. For each hypothesis we ran a Kruskal-Wallis test and Dunn post-hoc tests with median comparisons for each environment, gridworld and driving. For **H3**, Kruskal-Wallis did not yield statistically significant results for either gridworld or driving ($0.05 < p < 0.1$). Table 4 does show that our method achieves a higher mean user evaluation than the baselines; thus the lack of statistical significance could be due to a non-standardized evaluation scale. For **H4**, Kruskal-Wallis yielded statistically significant results for both environments ($H = 19, p < 0.0001$ for gridworld; $H = 14, p < 0.001$ for driving). Dunn further revealed that our method required a lower proportion of demonstrated states for gridworld ($p < 0.05$ for both baselines) and for driving ($p < 0.05$ for both baselines).

Our user study revealed two more interesting outcomes. First, we found that our method was much more sample efficient in the user study than in our empirical experiments. Comparing with the empirical results for an nEVD bound threshold of 0.3, our method required over 60% fewer demonstrations to be shown in the user study. We hypothesize that this is because the actual human demonstrators were more likely to choose highly informative demonstrations instead of random ones, enabling faster learning. Future work should investigate how close these human demonstrations are to optimally pedagogic demonstrations [18, 16]. Second, unlike in our empirical experiments, the user-provided demonstrations were indeed suboptimal at times; on average, 14% of user demos were suboptimal for gridworld, 8% for driving. The noise for driving shows that users aren't perfect at following even their own specified reward function, an interesting area future work can explore. Nevertheless, our approach still was able to efficiently and accurately determine demonstration sufficiency, indicating its robustness to noisy, real-world data.

## 6 DISCUSSION

*6.1 Pathway to Deployment.* Deploying our demonstration sufficiency methods onto a physical robot or other AI system is a matter of integrating the algorithms into or extending the robot or system's existing software and then having a human available to provide demonstrations. Demonstrations for physical robots are often provided through teleoperation, kinesthetic teaching, or even videos. Our methods assume that the robot shares the same features as the demonstrator, can correctly map demonstrated states and actions into its own state and action spaces, and can perform policy optimization (either model-based or model-free). While these are strong assumptions, they are common in HRI and are not unrealistic given recent advances in feature alignment [10, 9], cross-embodiment IRL [46], and offline reward and policy learning [41]. The exact hyperparameters used to implement our methods will depend on the risk-sensitivity of the environment and user discretion. The nEVD stopping condition can be used when no baseline policy exists or is able to be provided, or when the demonstrator wants to ensure confidence in minimizing policy loss itself. Meanwhile, the percent improvement stopping condition can be used in situations in which a baseline policy can be provided and the demonstrator is focused on improving this existing policy. Selecting the thresholds for the stopping conditions and the $\alpha$ value is also up to the demonstrator, though $\alpha = 0.95$ is most commonly used. Finetuning the thresholds and $\alpha$ will enable the demonstrator to adjust the robot's performance and conservativeness.

*6.2 Limitations and Future Work.* One of the limitations in our experiments is the repeated running of MCMC in the BIRL algorithm, which is time- and resource-intensive, especially as the number of samples increases. Implementing successor features could optimize transfer learning between different $R_{\text{MAP}}$ reward functions [6], improving MCMC efficiency. In addition, future work should explore the benefits of active queries in continuous-state domains.

Furthermore, while our empirical experiments and user study provide some evidence that our methodologies are compatible with suboptimal demonstrations, future work could make this application more robust by running a calibration stage before demonstration collection to estimate the suboptimality of the demonstrator [40, 21] and tune $\beta$ in the Bayesian inference algorithm. Finally, future work should study whether mutual information or posterior entropy could be used for estimating demonstration sufficiency.

*6.3 Conclusion.* In this paper, we formalized the problem of demonstration sufficiency and proposed and evaluated several methods which a robot can use to determine whether it has enough demonstration data. Our empirical and user study results provide promising evidence that our methods allow robots to self-assess their performance in cases where the reward function is unobserved by estimating this reward from human demonstrations. By developing robotic systems that can self-assess demonstration sufficiency, researchers and practitioners can achieve safer and more efficient training and deployment of LfD systems. Rather than simply giving robots as many demonstrations as possible and hoping that they will eventually learn the correct policy, our work takes the onus off the demonstrator by enabling robots and other AI systems to detect themselves when they are highly confident that they can use the existing demonstrations to learn a high-performing policy.

# Appendix

## A  CODE

Methods and experiments can be found here.

## B  METHOD IMPLEMENTATION DETAILS

Bayesian IRL (BIRL) uses Markov Chain Monte Carlo (MCMC) sampling in order to sample from the posterior $P(R|D)$ [38]. In this paper, we assume the prior distribution is uniform, though this distribution can take on any form depending on the domain. Feature weights $\hat{w}$ are sampled according to a proposal distribution and normalized such that $\|\hat{w}\|_2 = 1$. We implemented an adaptive version of MCMC where the standard deviation, or step size, $\sigma$ of the proposal distribution is automatically tuned during MCMC. If the current accept rate $r$ is higher than the target accept rate $r^*$, the step size will decrease by $\Delta\sigma$, and if it is lower, the step size will increase by $\Delta\sigma$, where $\Delta\sigma = \frac{\sigma}{\sqrt{i+1}}(r - r^*)$ and $i$ is the index of the current MCMC sample or iteration.

In our experiments we give the agent one demonstration, i.e., one state-action pair, at a time.

Algorithm 1 below shows a succinct pseudocode of our high-confidence nEVD bounding method. In particular, $N$, $\alpha$, and $\delta$ are prefixed values for the number of reward samples, the VaR quantile level and the confidence parameter, respectively. $F_{\mathcal{N}}^{-1}$ is the inverse Gaussian distribution.

---

**Algorithm 1:** Demonstration Sufficiency (nEVD)

1  Calculate the $\alpha$-VaR bound index $j$ with
   $j = \lceil N\alpha + F_{\mathcal{N}}^{-1}(1-\delta)\sqrt{N\alpha(1-\alpha)} - \frac{1}{2} \rceil$.
2  **for** $k = 0, 1, 2, \ldots$ **do**
3      Collect a new demo, update the posterior of the distribution of rewards $R$ by $\mathbb{P}(R|D) \propto \mathbb{P}(D|R)\mathbb{P}(R)$.
4      Using MCMC, compute $R_{\text{MAP}}$ and randomly sample rewards $R_1, R_2, \ldots, R_N$.
5      Run value iteration on $R_{\text{MAP}}$ and extract policy $\pi_{\text{MAP}} =: \pi_{\text{robot}}$.
6      Perform policy evaluation of $\pi_{\text{robot}}$ on each reward sample $R_i$ to obtain $V_{R_i}^{\text{robot}}$.
7      Perform policy evaluation of $\pi_{\text{rand}}$ on each reward sample $R_i$ to obtain $V_{R_i}^{\text{rand}}$.
8      Run value iteration on each $R_i$ to obtain $V_{R_i}^*$.
9      Calculate $\text{nEVD}_i = \frac{V_{R_i}^* - V_{R_i}^{\text{robot}}}{V_{R_i}^* - V_{R_i}^{\text{rand}}}$. Sort and find the $\alpha$-VaR bound $\text{nEVD}_j$.
10     **if** $nEVD_j > threshold$ **then**
11         Repeat.
12     **else**
13         Stop.
14     **end**
15 **end**

---

## C  USER STUDY DETAILS

Here we provide a more thorough description of our user study setup. Users were asked to complete six rounds (3 demonstration sufficiency methods × 2 environments), without knowing which round used which method. For each round, they were presented with either a gridworld or driving environment to teach the agent in. Users were instructed to sequentially provide demonstrations, which were (state, action) pairs, via the online interface, until the robot declared demonstration sufficiency.

For gridworld, users were shown a reward function as a weight vector, where each reward value was color-coded to match a feature. They were told to guide the robot towards the goal as fast as possible while avoiding low-reward features. For driving, we described the features and requested users create their own reward function as a weighted combination of those features (three lanes, collision, and dirt patch). We provided examples to help, such as, "If you want to drive towards the right as much as possible and avoid accidents, your reward function could be 1, 2, 3, -10, -5." Their reward function was then normalized to have an L2 norm of 1 to be consistent with our methodology. Users were then told to give demonstrations according to this custom reward function.

At the end of each of the six rounds, users were shown a visual display of the robot's learned policy and asked, "On a scale of 1 (worst) to 5 (best), how well did the agent's learned policy match your intended policy or reward function?"

A short video example of what a user study round would have looked like can be found here.

## D  REASONING UNDER UNCERTAINTY

A particular benefit to using a Bayesian approach for determining demonstration sufficiency is that this allows agents to properly reason about their uncertainties after receiving some number of demonstrations. In the paper we analyzed how our method can still perform strongly when noisy or suboptimal demonstrations are provided. Here we offer additional insight on how it performs with ambiguous (but still optimal) demonstrations. We define an ambiguous demonstration as one that is either redundant—one that has been provided before and thus gives no new information—or unhelpful—one that is unclear at showing what objectives or obstacles are present in the environment, in other words one that could be applicable to a wider set of environment configurations. Given such ambiguous demonstrations, we find that an agent using our approach will still be able to identify that there remains a high level of uncertainty about the demonstrator's intent, and so it will request additional demonstrations. On the other hand, when the demonstrations are clear, the agent can determine they are sufficient early on.

Figure 5 shows examples of ambiguous vs. informative demonstrations for each of the four environments, as well as results from a small experiment. We used the nEVD stopping condition with a threshold of 0.5 and provided each type of demonstration separately; in other words, we gave either only very informative or only very ambiguous demonstrations per trial, though it is important to note that in practice, there will generally be a mix of the two types. Our results show that the more ambiguous the demonstrations provided, the more additional demonstrations the agent using our Bayesian approach will request, indicating its ability to correctly maintain a level of uncertainty about whether or not it will be able to learn the correct policy.

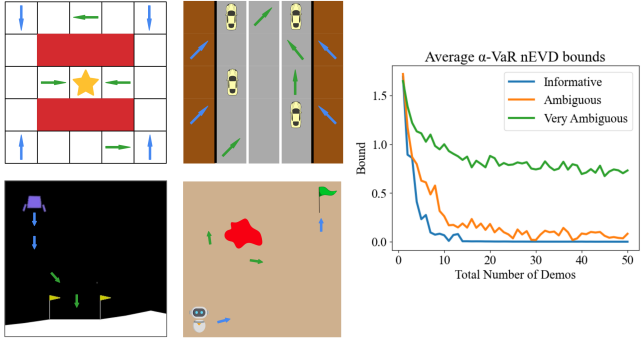

**Figure 5: Comparison of number of demonstrations requested. Left: Sample environments in which we provided ambiguous (blue arrows) vs. informative (green arrows) demonstrations. Right: This plot illustrates that the more ambiguous the provided demonstrations, the more the agent maintains uncertainty and requests additional demos. The more informative the demonstrations already are, the fewer demos the agent will request.**

## E  COMPLETE RESULTS

Below we show complete results for our proposed stopping conditions using both passive and active demonstration selection as well as baseline stopping methods. In addition to the dependent measures described in section 4.3, we also show

(1) Bound error: the discrepancy between the robot's final $\alpha$-VaR bound and the true nEVD or percent improvement value
(2) Policy optimality: the proportion of actions under $\pi_{\text{robot}}$ that will lead to an optimal Q-state
(3) Accuracy: for the nEVD stopping condition, the proportion of MDP replicates in which the bound is correctly greater than or equal to the ground-truth nEVD; for the percent improvement stopping condition, the proportion of replicates in which the bound is correctly less than or equal to the ground-truth percent improvement

### E.1  nEVD Stopping Condition

Figures 6 and 7 are the results of testing the nEVD stopping condition. Figure 6 shows the results when using passive demonstration selection. Figure 7 shows the results when using active query demonstration selection.

### E.2  Baseline Stopping Condition Results

Figure 8 is the result of testing the patience baseline. Figure 9 is the result of testing the held-out set baseline.

### E.3  Percent Improvement Stopping Condition

Figures 10 and 11 are the results of testing the percent improvement stopping condition. The $x$-axis displays the possible thresholds on percent improvement. Figure 10 shows the results when using passive demonstration selection. Figure 11 shows the results when using active query demonstration selection.

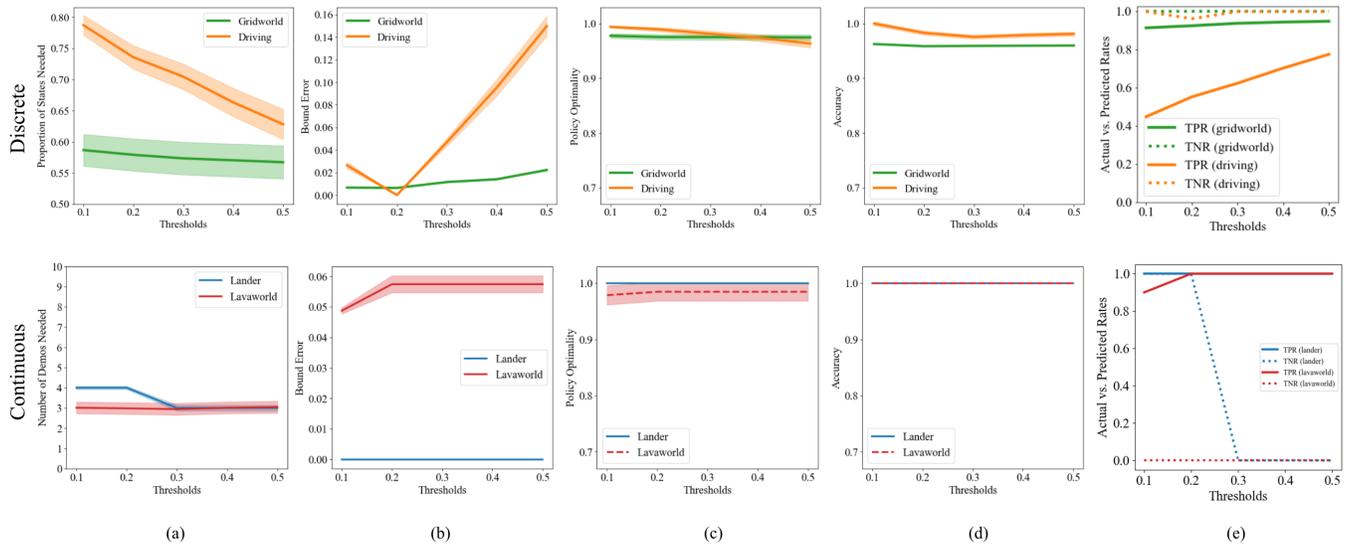

Figure 6: nEVD stopping condition (passive selection): Error bands denote standard error. Moving left to right across the $x$-axis denotes higher thresholds, i.e. looser requirements for the $\alpha$-VaR bound. The looser the threshold, the fewer states need to be seen before the robot determines demonstration sufficiency but the higher the $\alpha$-VaR bound error as well, as seen in columns (a) and (b). Policy optimality and accuracy, columns (c) and (d), stay high across all thresholds due to the nature of the $\alpha$-VaR bound. The high true positive rates in column (e) indicate the robot is able to learn a policy such that its ground-truth nEVD is *also* below the threshold.

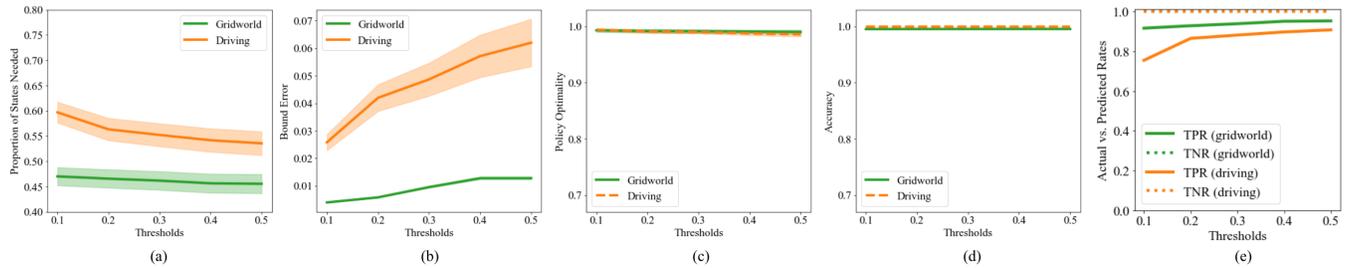

Figure 7: nEVD stopping condition (active selection): The same patterns for the above experiment hold true.

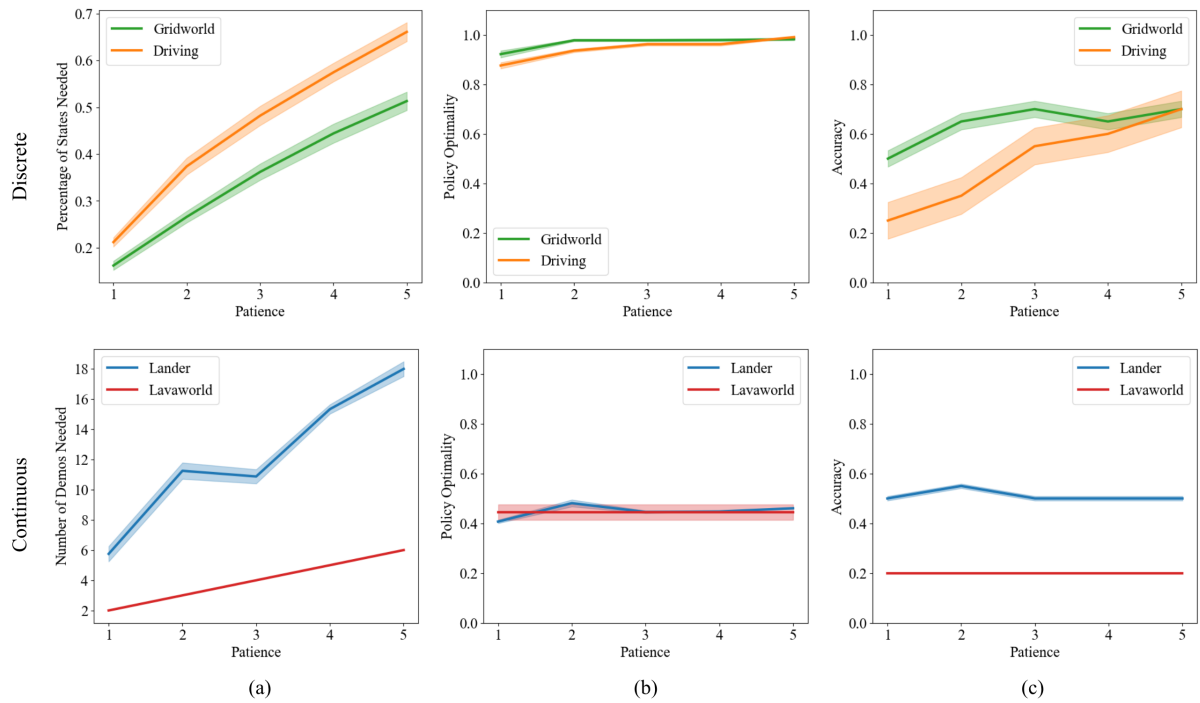

Figure 8: Patience stopping condition: Moving across the $x$-axis denotes higher patience, i.e. more iterations of demonstrations in which the robot is expected to reach the same MAP policy. Column (a) shows the low number of demonstrations needed, especially for lower patience. While policy optimality in column (b) is acceptable, column (c) shows that accuracy is quite low and variable. Error bands denote standard error.

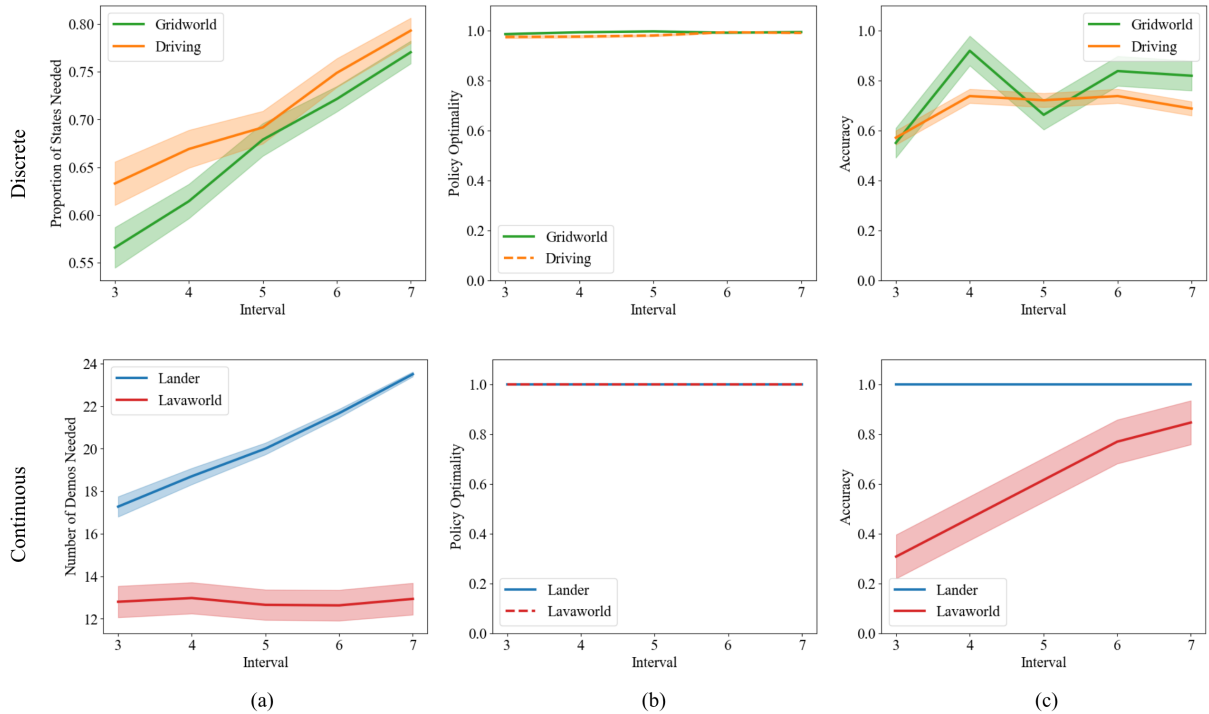

Figure 9: Held-out set baseline: Moving across the $x$-axis denotes larger intervals, i.e. demonstrations are added to the held-out set more infrequently. Column (a) shows the extremely high number of demonstrations needed, especially in the continuous domain. While policy optimality in column (b) is still high, this comes at the cost of waiting for multiple demonstrations to be added to a validation set and for the validation set to be of sufficient size for comparison. Column (c) shows that accuracy is quite low and variable. Again, error bands denote standard error.

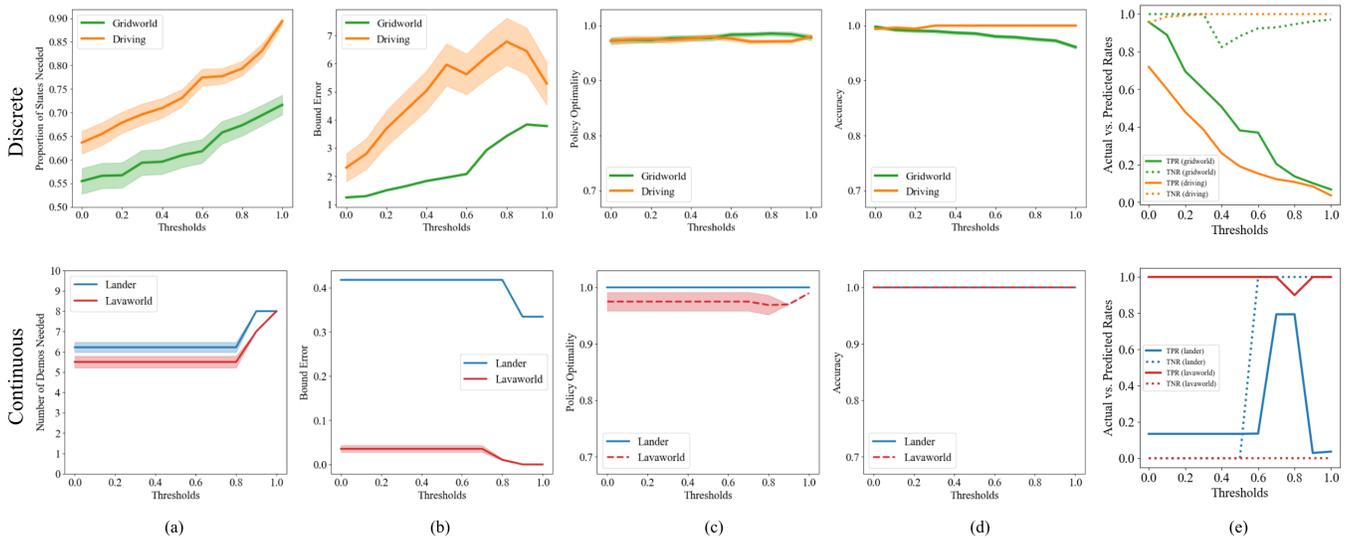

Figure 10: Percent improvement stopping condition (passive selection): Error bands denote standard error. Moving left to right on the $x$-axis shows higher thresholds, i.e. stricter requirements for the percent improvement over the baseline policy. As seen in columns (a) and (b), the stricter the requirement, the more states will need to be seen and the higher the bound error will reach, as the robot becomes more conservative. Columns (c) and (d) still show high policy optimality and accuracy. Column (e)'s lower TPR highlights 1) the agent's conservativeness and 2) the relatively more difficult feat of also achieving near 100% improvement—performing twice as well as the baseline—under the true reward function.

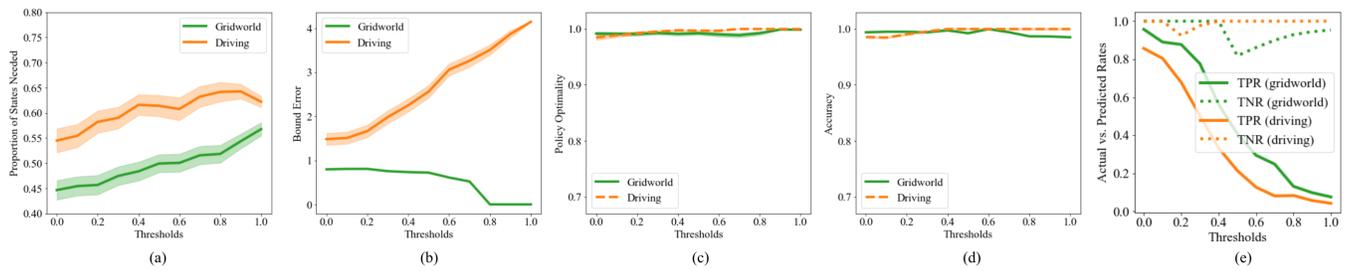

Figure 11: Percent improvement stopping condition (active selection): The same patterns for the above experiment hold true.